\documentclass[runningheads,a4paper]{llncs}

\usepackage{amssymb}
\usepackage{graphicx}
\usepackage{array}
\newcolumntype{L}[1]{>{\raggedright\let\newline\\\arraybackslash\hspace{0pt}}m{#1}}
\newcolumntype{C}[1]{>{\centering\let\newline\\\arraybackslash\hspace{0pt}}m{#1}}
\newcolumntype{R}[1]{>{\raggedleft\let\newline\\\arraybackslash\hspace{0pt}}m{#1}}
\usepackage[belowskip=-15pt,aboveskip=0pt]{caption}

\usepackage[colorinlistoftodos]{todonotes}
\usepackage{enumitem}

\usepackage{url}
\urldef{\mailsa}\path|{p.t.j.m.vossen, s.baezsantamaria, l.bajcetic, b.kraaijeveld}@vu.nl|   
\newcommand{\keywords}[1]{\par\addvspace\baselineskip
\noindent\keywordname\enspace\ignorespaces#1}

\begin{document}

\mainmatter  % start of an individual contribution

% first the title is needed
\title{Leolani: a reference machine with a theory of mind for social communication}

% a short form should be given in case it is too long for the running head
\titlerunning{Leolani: a reference machine with a theory of mind for social communication}

% the name(s) of the author(s) follow(s) next.
%
\author{Piek Vossen%
\and Selene Baez \and Lenka Baj\u{c}eti\'{c}  \and Bram Kraaijeveld}
\authorrunning{Piek Vossen%
\and Selene Baez \and Lenka Baj\u{c}eti\'{c}  \and Bram Kraaijeveld}
% (feature abused for this document to repeat the title also on left hand pages)

% the affiliations are given next; don't give your e-mail address
% unless you accept that it will be published
\institute{VU University Amsterdam, Computational Lexicology and Terminology Lab, De Boelelaan 1105, 1081HV Amsterdam, The Netherlands\\
\mailsa\\
\url{www.cltl.nl}}

\maketitle

%%% Stilistic Conventions %%%
% 1. Do use the Oxford comma
% 2. Citations between parenthesis and separated by a comma
% 3. Picture should be after you mention it

\vspace{-10pt}
\begin{abstract}
Our state of mind is based on experiences and what other people tell us. This may result in conflicting information, uncertainty, and alternative facts. We present a robot that models relativity of knowledge and perception within social interaction following principles of the \textit{theory of mind}. We utilized vision and speech capabilities on a Pepper robot to build an interaction model that stores the interpretations of perceptions and conversations in combination with provenance on its sources. The robot learns directly from what people tell it, possibly in relation to its perception. We demonstrate how the robot's communication is driven by hunger to acquire more knowledge from and on people and objects, to resolve uncertainties and conflicts, and to share awareness of the perceived environment. Likewise, the robot can make reference to the world and its knowledge about the world and the encounters with people that yielded this knowledge.
%DONE{reduced to 149}

\end{abstract}
% https://www.tsdconference.org/tsd2018/paper_instr.html
%the deadline is May 31 and the format is Springer LNCS approx. 8 pages
\vspace{-20pt}
\section{Introduction}
\label{sec:intro}
%\vspace{-10pt}
%the deadline is May 31 and the format is Springer LNCS approx. 8 pages
%(see https://www.tsdconference.org/tsd2018/paper_instr.html for templates).
% Motivation
People make mistakes; but machines err as well \cite{mirnig2017err} as there is no such thing as a perfect machine. Humans and machines should therefore recognize and communicate their ``imperfectness" when they collaborate, especially in case of robots that share our physical space. Do these robots perceive the world in the same way as we do and, if not, how does that influence our communication with them? How does a robot perceive us? Can a robot trust its own perception? Can it believe and trust what humans claim to see and believe about the world? For example, if a child gets injured, should a robot trust their judgment of the situation, or should it trust its own perception?  How serious is the injury, how much knowledge does the child have, and how urgent is the situation? How different would the communication be with a professional doctor?

% Participants in a conversation matter just as much as the purpose of it
Human-robot communication should serve a purpose, even if it is just (social) chatting. Yet, effective communication is not only driven by its purpose, but also by the communication partners and the degree to which they perceive the same things, have a common understanding and agreement, and  trust. One of the main challenges to address in human-robot communication is therefore to handle uncertainty and conflicting information. We address these challenges through an interaction model for a humanoid-robot based on the notion of a `theory of mind'  (\cite{premackwoodruff1978,leslie1987pretense}). The `theory of mind' concept states that children at some stage of their development become aware that other people's knowledge, beliefs, and perceptions may be untrue and/or different from theirs. Scassellati (\cite{scassellati2001foundations} and \cite{scassellati2002theory}) was the first to argue that humanoid robots should also have such an awareness. We take his work as a starting point for implementing these principles in a Pepper robot, in order to drive social interaction and communication. 

Our implementation of the theory of mind heavily relies on the Grounded Representation and Source Perspective model (GRaSP) (\cite{vossen2016kbs}, \cite{fokkens2017grasp}). GRaSP is an RDF model representing situational information about the world in combination with the perspective of the sources of that information. The robot brain does not only record the knowledge and information as symbolic interpretations, but it also records from whom or through what sensory signal it was obtained. The robot acquires knowledge and information both from the sensory input as well as directly from what people tell it. The conversations can have any topic or purpose but are driven by the robot's need to resolve conflicts and ambiguities, to fill gaps, and to obtain evidence in case of uncertainty. %Currently the robot is designed to have social conversations but this can be adapted to any other purpose.

% Structure
This paper is structured as follows. Section \ref{sec:related_work} briefly discusses related work on theory of mind and social communication. In Section \ref{sec:grasp}, we explain how we make use of the GRaSP model to represent a theory of mind for the robot. Next, Section \ref{sec:model} describes the implementation of the interaction model built on a Pepper robot. 
%Section \ref{sec:technical} provides the technical specifications of the components in the model. 
Finally, Section \ref{sec:future_work} outlines the next steps for improvement and explores other possible extensions to our model. We list a few examples of conversations and information gathered by the robot in the Appendix. 

\vspace{-5pt}
\section{Related Work}
\label{sec:related_work}
\vspace{-5pt}

%\textit{}\vspace{-10pt}
%Scassellati (\cite{scassellati2001foundations} and \cite{scassellati2002theory}) argues that social humanoid robots need a model of the 'theory of mind' for social communication. 
Theory of mind is a cognitive skill to correctly attribute beliefs, goals, and percepts to other people, and is assumed to be essential for social interaction and for the development of children \cite{leslie1987pretense}. The theory of mind allows the truth properties of a statement to be based on mental states rather than observable stimuli, and it is a required system for understanding that others hold beliefs that differ from our own or from the observable world, for understanding different perceptual perspectives, and for understanding pretense and pretending. Following \cite{baron1997mindblindness}, Scassellati decomposes this skill into stimuli processors that can detect static objects (possibly inanimate), moving objects (possibly animate), and objects with eyes (possibly having a mind) that can gaze or not (eye-contact), and a shared-attention mechanism to determine that both look at the same objects in the environment. His work further focuses on the implementation of the visual sensory-motor skills for a robot to mimic the basic functions for object, eye-direction and gaze detection. He does not address human communication, nor the storage of the result of the signal processing and communication in a brain that captures a theory of mind. In our work, we rely on other technology to deal with the sensory data processing, and add language communication and the storage of perceptual and communicated information to reflect conflicts, uncertainty, errors, gaps, and beliefs. 

More recent work on the 'theory of mind' principle for robotics appears to focus on the view point of the human participant rather than the robot's. These studies reflect on the phenomenon of anthropomorphism \cite{ono2000reading} \cite{epley2007seeing}: the human tendency to project human attributes to nonhuman agents such as robots. Closer to our work comes \cite{hiatt2011accommodating} who use the notion of a theory of mind to deal with human variation in response. The robot runs a simulation analysis to estimate the cause of variable behaviour of humans and likewise adapts the response. However, they do not deal with the representation and preservation of conflicting states in the robot's brain. To the best of our knowledge, we are the first that complement the pioneering work of Scassellati with further components for an explicit model of the theory of mind for robots (see also \cite{mavridis2015review} for a recent overview of the state-of-the-art for human-robot interactive communication).

There is a long-tradition of research on multimodal communication \cite{partan2005issues}, human-computer-interfacing \cite{card2017psychology}, and other component technologies such as face detection \cite{viola2004robust}, facial expression, and gesture detection \cite{kanade2000comprehensive}. The same can be said about multimodal dialogue systems \cite{wahlster2006smartkom}, and more recently, around chatbot systems using neural networks \cite{serban2016building}. In all these studies the assumption is made that systems process signals correctly, and that these signals can be trusted (although they can be ambiguous or underspecified). In this paper, we do not address these topics and technologies but we take them as given and focus instead on the fact that they can result in conflicting information, information that cannot be trusted or that is incomplete within a framework of the theory of mind. Furthermore, there are few systems that combine natural language communication and perception to combine the result in a coherent model. An example of such work is \cite{she2017interactive} who describe a system for training a robot arm through a dialogue to perform physical actions, where the ``arm" needs to map the abstract instruction to the physical space, detect the configuration of objects in that space, determine the goal of the instructions. Although their system deals with uncertainties of perceived sensor data and the interpretation of the instructions, it does not deal with modeling long-term knowledge, but only stores the situational knowledge during training and the capacity to learn the action. As such, they do not deal with conflicting information coming from different sources over time or obtained during different sessions. Furthermore, their model is limited to physical actions and the artificial world of a few objects and configurations. 

%\cite{mirnig2017err} force errors on the robot behaviour to study the way people respond.  Participants had to rate the robot?s anthropomorphism, likability, and perceived intelligence. Their results show that participants liked the faulty robot significantly better than the robot that interacted flawlessly. They do not consider natural technical errors that robots commonly make and they do not adapt the robot behaviour on the responses of the humans on these errors. As such they do not model collaborative behaviour.

%paper on user-centric conversation for the Alexa price \cite{2018arXiv180410202F}. The ChatBot, designed to participate in the 2017 Amazon Alexa Prize, is user-centric and content-drieven. User-centric because users control the topic, and the systems adapts answers to the user's interest by gauging the user's personality. It is content-driven because it continuously supplies 'interesting' and 'relevant' information. A knowledge-graph is updated on a daily basis from existing external knowledge bases. Specialised conversation agents, so-called miniskills, access the knowledge to push it to the users. The model does not allow pushing knowledge to the system and cannot model and handle conflicts. %Quote from \cite{2018arXiv180410202F}. "speaker?s intent or goals, the desired topic or potential subtopics of conversation, and the stance or sentiment of a user?s reaction to a system comment."

\vspace{-5pt}
\section{GRaSP to model the theory of mind}
\label{sec:grasp}
\vspace{-5pt}

The main challenges for acquiring a theory of mind is the storage of the result of perception and communication in a single model, and the handling of uncertainty and conflicting information.
%: how should the robot deal with flawed output from the speech recognition and computer vision components, ambiguous or vague instructions, or conflicting information provided by different speakers and perceptions?
We addressed these challenges by explicitly representing all information and observations processed by the robot in an artificial brain (a triple store) using the GRaSP model \cite{fokkens2017grasp}. 
%This RDF model represents facts and situational information about the world in combination with the perspective of the sources of that information. The robot brain not only records the knowledge and information as symbolic interpretations but it also records from whom or through what sensory signal it was obtained. The perspectives of these sources range from source certainty, past-present-future temporal anchoring, source sentiment, emotion and judgment, source epistemics (believe, denial) and it may reflect various strength. The robot acquires knowledge and information both from the sensory input as well as directly from what people tell it. 
For modeling the interpretation of the world, GRaSP relies on the Simple Event Model (SEM) \cite{van2011design} an RDF model for representing instances of events. RDF triples are used to relate event instances with \textit{sem:hasActor}, \textit{sem:hasPlace} and \textit{sem:hasTime} object properties to actors, places, and time also represented as resources. For example, the triples [laugh, sem:hasActor, Bram], [laugh, sem:hasTime, 20180512] express that there was a \textit{laugh} event involving \textit{Bram} on the \textit{12th of May 2018}. 

GRaSP extends this model with \textit{grasp:denotedIn} links to express that the instances and relations in SEM have been mentioned in a specific signal, e.g. a camera signal,  human speech, written news. These signals are represented as \textit{grasp:Chat} and \textit{grasp:Turn} which: a) are subtypes of \textit{sem:Event} and therefore linked to an actor and time, and b) derive \textit{grasp:Mention} objects which point to specific mentioning of entities and events in the signal. Thus, if \textit{Lenka} told the robot ``Bram is laughing", then this expression is considered as a speech signal that mentions the entity \textit{Bram} and the event instance \textit{laugh}, while the time of the utterance is given and correlates with the tense of the utterance. 

\begin{table}[!ht]
\vspace{-20pt}
{\scriptsize
\begin{tabular}{L{4.1cm}L{3cm}L{4.6cm}} 
leolaniWorld:instances & & \\
\hline
  leolaniWorld:Lenka    &  rdfs:label    &       ``Lenka";\\
  leolaniWorld:Bram    & rdfs:label     &      ``Bram";\\
      & grasp:denotedIn    & leolaniTalk:chat1\_turn1\_char0-16.\\
\hline
  leolaniWorld:laugh  & a & sem:Event;\\
   &   rdfs:label  & ``laugh"; \\
   &   grasp:denotedIn     &  leolaniTalk:chat1\_turn1\_char0-16.\\
\hline
\end{tabular}
\label{rdf:world}
}
\vspace{-20pt}
\end{table}

\noindent GRaSP further allows to express properties of the mentions such as the source (using \textit{prov:wasAttributedTo}\footnote{Where possible, we follow the PROV-O model: \url{https://www.w3.org/TR/prov-o/}}), and the perspective of the source towards the content or claim (using \textit{grasp:hasAttribution}). In the case of robot interactions, the source of a spoken utterance is the person identified by the robot, represented as a \textit{sem:Actor}. Finally, we use \textit{grasp:Attribution} to store properties related to the perspective of the source to the claimed content of the utterance: what emotion is expressed, how certain is the source, and/or if the source confirms or denies it.  Following this example, the utterance is attributed to \textit{Lenka}; thus we model that Lenka \textit{confirms} Bram's laughing, and that she is \textit{uncertain} and \textit{surprised}. The perspective subgraph resulting from the conversation would look as follows: 

\begin{table}[!ht]
\vspace{-15pt}
{\scriptsize
\begin{tabular}{L{4.1cm}L{3cm}L{4.6cm}} 
leolaniTalk:perspectives & & \\
\hline
leolaniTalk:chat1\_turn1    & a & grasp:Turn;\\
    &   sem:hasActor   & leolaniFriends:Lenka;\\
    &   sem:hasTime       &        leolaniTime:20180512.\\
\hline
leolaniTalk:chat1\_turn1\_char0-16 & a  & grasp:Mention; \\
     & grasp:denotes  &  leolaniWorld:claim1 ;\\
     & prov:wasDerivedFrom  &  leolaniTalk:chat1\_turn1 ;\\
     & prov:wasAttributedTo  &  leolaniFriends:Lenka .\\
\hline
\multicolumn{2}{l}{leolaniTalk:chat1\_turn1\_char0-16\_ATTR1 	a} & grasp:Attribution; \\   
     & rdf:value          &              grasp:CONFIRM, grasp:UNCERTAIN, grasp:SURPRISE;\\
     & grasp:isAttributionFor    & leolaniTalk:chat1\_turn1\_char0-16.\\
\hline

\end{tabular}
\label{rdf:perspective}
}
\vspace{-26pt}
\end{table}

\noindent Our model represents the claims containing the SEM event and its relations as:

\begin{table}[!ht]
\vspace{-20pt}
{\scriptsize
\begin{tabular}{L{4.2cm}L{3cm}L{4.5cm}}  
leolaniWorld:claims & & \\
\hline
leolaniWorld:claim1  &    a       &     grasp:Statement;\\
  &    grasp:subject       &     leolaniWorld:laugh;\\
  &    grasp:predicate    &     sem:hasActor;\\
  &    grasp:object         &     leolaniFriends:Bram.\\
\hline 
\end{tabular}
\label{rdf:claims}
}
\vspace{-10pt}
\end{table}

%GRaSP:subject predicate object do not exist ... do we still need claims if we are now event centric?

\noindent Now assume that \textit{Selene} is also present and she denies that Bram is laughing by saying: ``No, Bram is not laughing". This utterance then gets a unique identifier e.g. \textit{leolaniTalk:chat2\_turn1}, while our Natural Language processing will derive exactly the same claim as before. The only added information is therefore the mentioning of this claim by \textit{Selene} and her perspective, expressed as:

\begin{table}[!ht]
\vspace{-15pt}
{\scriptsize
\begin{tabular}{L{4.1cm}L{3cm}L{4.6cm}} 
leolaniTalk:perspectives & & \\
\hline
leolaniTalk:chat2\_turn1    & a & grasp:Turn;\\
    &   sem:hasActor   & leolaniFriends:Selene;\\
    &   sem:hasTime       &        leolaniTime:20180512.\\
\hline
leolaniTalk:chat2\_turn1\_char0-24 & a  & grasp:Mention; \\
     & grasp:denotes  &  leolaniWorld:claim1 .\\
     & prov:wasDerivedFrom  &  leolaniTalk:chat2\_turn1 .\\
     & prov:wasAttributedTo  &  leolaniFriends:Selene .\\
\hline
\multicolumn{2}{l}{leolaniTalk:chat2\_turn1\_char0-24\_ATTR1	a} & grasp:Attribution; \\   
     & rdf:value          &              grasp:DENY, grasp:CERTAIN;\\
     & grasp:isAttributionFor    & leolaniTalk:chat2\_turn1\_char0-24.\\
\hline

\end{tabular}
\label{rdf:turnmary}
}
\vspace{-20pt}
\end{table}

\noindent Along the same lines, if \textit{Lenka} now agrees with \textit{Selene} by saying ``Yes, you are right", we model this by adding only another utterance of \textit{Lenka} and her revised perspective to the same claim, as shown below.\footnote{There are now two perspectives from \textit{Lenka} on the same claim (she changed his mind), expressed in two different utterances} 

\begin{table}[!ht]
\vspace{-15pt}
{\scriptsize
\begin{tabular}{L{4.1cm}L{3cm}L{4.6cm}} 
leolaniTalk:perspectives & & \\
\hline
leolaniTalk:chat1\_turn2    & a & grasp:Turn;\\
    &   sem:hasActor   & leolaniFriends:Lenka;\\
    &   sem:hasTime       &        leolaniTime:20180512.\\
\hline
leolaniTalk:chat1\_turn2\_char0-18 & a  & grasp:Mention; \\
     & grasp:denotes  &  leolaniWorld:claim1 .\\
     & prov:wasDerivedFrom  &  leolaniTalk:chat1\_turn2 .\\
     & prov:wasAttributedTo  &  leolaniFriends:Lenka .\\
\hline
\multicolumn{2}{l}{leolaniTalk:chat1\_turn2\_char0-18\_ATTR2	a} & grasp:Attribution; \\   
     & rdf:value          &   grasp:DENY, grasp:CERTAIN;\\
     & grasp:isAttributionFor    & leolaniTalk:chat1\_turn2\_char0-18.\\
\hline

\end{tabular}
\label{rdf:turnbill2}
}
\vspace{-20pt}
\end{table}

In the above examples, we only showed information given to the robot through conversation. GRaSP can however deal with any signal and  we can therefore also represent sensor perceptions as making reference to the world or people that the robot knows. Assuming that the robot also sees and recognizes \textit{Bram}, about whom \textit{Lenka} and \textit{Selene} are talking, this can be represented as follows, where we now include all the other mentions from the previous conversations:

\begin{table}[!ht]
\vspace{-15pt}
{\scriptsize
\begin{tabular}{L{4.1cm}L{3cm}L{4.6cm}} 
leolaniWorld:instances & & \\
\hline
leolaniWorld:Bram    & rdfs:label     &      ``Bram";\\
      & grasp:denotedIn    & leolaniTalk:chat1\_turn1\_char0-16,\\
&        & leolaniTalk:chat2\_turn1\_char0-24,\\
&       & leolaniTalk:chat1\_turn2\_char0-18;\\
&      grasp:denotedBy  & leolaniSensor:FaceRecognition1.\\
\hline
\end{tabular}
\label{rdf:sensor}
}
\vspace{-20pt}
\end{table}

\noindent A facial expression detection system could detect Bram's emotion and store this as perspective by the robot, e.g. [leolaniSensor:FaceRecognition1, rdf:value, grasp:SAD], in addition to \textit{Lenka} and \textit{Selene} on Bram's state of mind.

As all data are represented as RDF triples, we can easily query all claims made and all properties stored by the robot on instances of the world. We can also query for all signals (utterances and sensor data) which mention these instances and all perspectives that are expressed. The model further allows to store certainty values for observations and claims as well as the result of emotion detection in addition to the content of utterances (e.g. through modules for facial expression detection or voice-emotion detection). Finally, all observations and claims can be combined with background knowledge on objects, places, and people available as linked open data (LOD).

Things observed by the robot in the environment and things mentioned in conversation are thus stored as unified data in a "brain" (triple store). This brain contains identified people with whom the robot communicates, perceived objects about which they communicated\footnote{The robot continuously detects objects, but these are only stored in memory when they are referenced by humans in the communication}, as well as properties identified or stated of these objects or people. Given this model, we can now design a robot communication model in combination with sensor processing on top of a theory of mind. In the next section, we explain how we implemented this model and what conversations can be held.

\vspace{-5pt}
\section{Communication model}
\label{sec:model}

\begin{figure}[h]
\vspace{-25pt}
\centering
  \includegraphics[width=\textwidth]{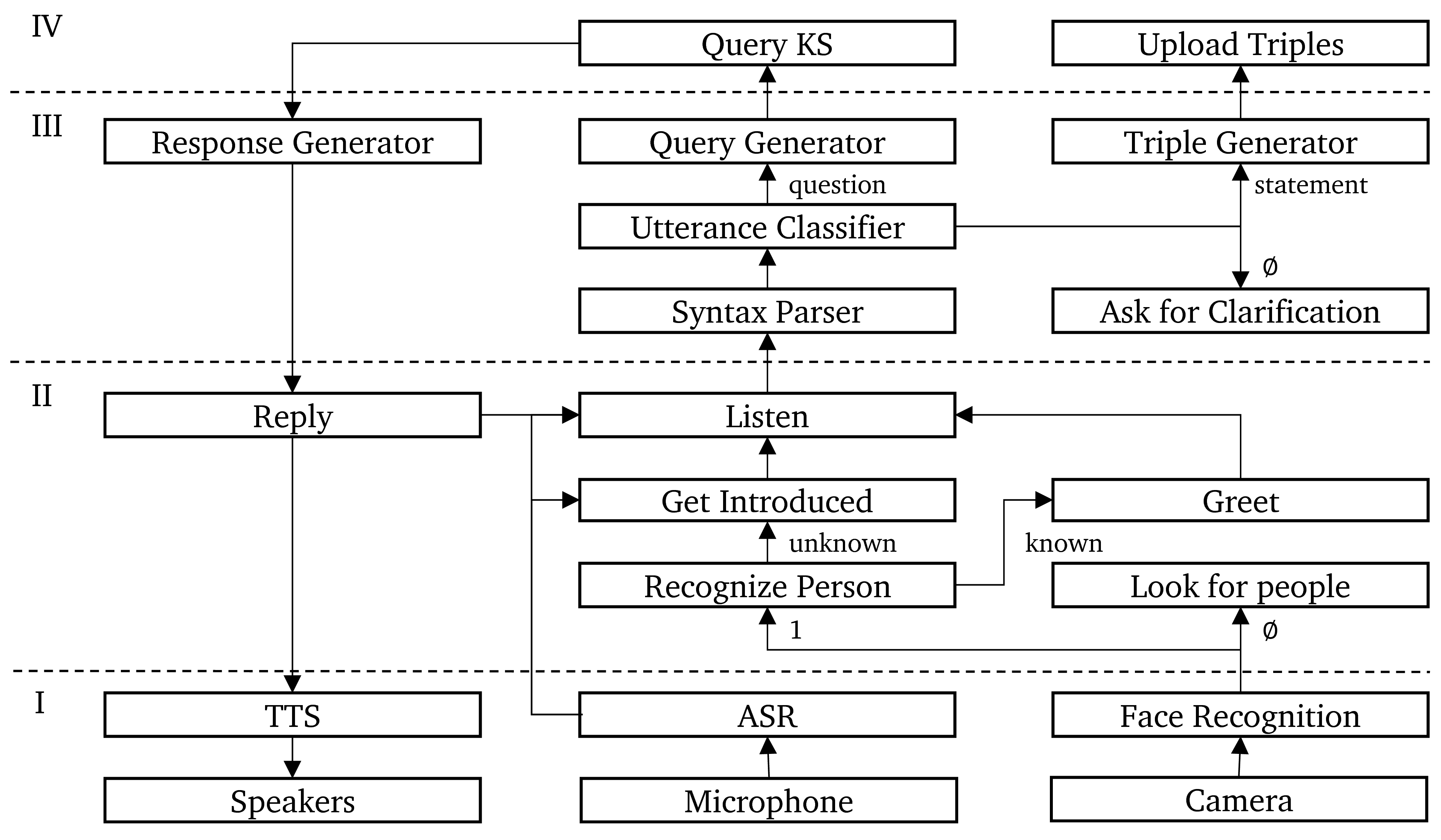}
  \caption{The four layer conversation model, comprised of I. Signal Processing layer, II. Conversation Flow layer, III. Natural Language layer and IV. Knowledge Representation layer.}
  \label{fig:four_layers}
  \vspace{-5pt}
\end{figure}
\noindent Our communication model consists of four layers: Signal Processing layer, Conversation Flow layer, Natural Language layer, and Knowledge Representation layer, which are summarized in Figure \ref{fig:four_layers}. Signal Processing (I) establishes the mode of input through which the robot acquires experiences (vision and sound) but also knowledge (communication). The Conversation Flow layer (II) acts as the controller, as it determines the communicative goals, how to interpret human input, and whether the robot should be proactive. Layer III is the Natural Language layer that processes utterances and generates expressions, both of which can be questions or statements. Incoming statements are stored in the brain, while questions are mapped to SPARQL queries to the brain. SPARQL queries are also initiated by the controller (layer II) on the basis of sensor data (e.g. recognizing a face or not) or the state of the brain (e.g. uncertainty, conflicts, gaps) without a human asking for it. The next subsections briefly describe the four layers. We illustrate the functions through example dialogues that are listed in the Appendix. Our robot has a name \textit{Leolani}, which is Hawaiain for \textit{voice of an angel}, and a female gender to make the conversations more natural. %No further character traits are implemented for the robot. 

\begin{enumerate}
\vspace{-10pt}
\item \textbf{Signal Processing}
Signal processing is used to give the robot awareness of its (social) surroundings and to recognize the recipient of a conversation. For assessing the context of a conversation, the robot has been equipped with eye contact detection, face detection, speech detection, and object recognition. These modules run continuously as the robot attempts to learn and recognize its surroundings. Speech detection is performed using WebRTC\cite{WebRTC} and object recognition has been built on top of the Inception\cite{inception} neural network through TensorFlow\cite{tensorflow2015-whitepaper}. During conversation the robot utilizes face recognition and speech recognition to understand who says what. Face recognition has been implemented using OpenFace\cite{amos2016openface} and speech recognition is powered by the Google Speech API\cite{GoogleSTT}.

\item \textbf{Conversation Flow}
In order to guide and respond during a one-to-one conversation, the robot needs to reason over its knowledge (about itself, the addressee, and the world) while taking into account its goals for the interaction. To model this we follow a Belief, Desire, Intention (BDI)\cite{bratman1987intention} approach. 

\underline{Desires}: The robot is designed to be hungry for social knowledge. This includes desires like asking for social personal information (name, profession, interests, etc.), or asking for knowledge to resolve uncertainties and conflicts.

\underline{Beliefs}: We consider the output of the other three layers to be part of the core beliefs of the robot, thus including information about what is being \textit{sensed}, \textit{understood}, and \textit{remembered} during a conversation. 

\underline{Intentions}: The set of current beliefs combined with the overarching desires then determine the next immediate action to be taken (aka. intention). The robot is equipped with a plan library including all possible intentions such as: a) Look for a person to talk to, b) Meet a new person, c) Greet a known person, d) Detect objects, and e) Converse (including sub-intentions like Ask a question, State a fact, Listen to talker, and Reply to a question). The dialogues in the Appendix illustrate this behavior.

\item \textbf{Natural Language}
During a conversation, information flows back and forth. Thus, one of the goals of this layer is to transform natural language into structured data as RDF triples. When the robot listens, the utterances are stored along with the information about their speaker. After the perceived speech is converted to text, it is tokenized. The NLP module first determines if the utterance is a question or statement, because these are parsed differently. The parser consists of separate modules for different types of words, which are called on-demand, thus not clogging the NLP pipeline unnecessarily. This is important as Leolani needs to analyze an utterance and respond fast in real-time. Currently, the classification of the roles of words, such as predicate, subject, object, is done by a rule-based system.
%as we want to mimic the way humans learns a second language, initially starting from a basic vocabulary. 
%In the future this can be changed, and we could still reuse the components for analyzing different roles. 
Next, the subject, object and predicate relations are mapped to triples for storage or querying the brain.
%Roles in this case are predicates, subjects or objects, but the number of roles will grow as the project develops.
This module also performs a perspective analysis over the utterance. Negation, certainty and sentiment or emotion are extracted separately from the text and their values added to the triple representation. A second goal of this layer is to \textit{produce} natural language, based on output from either layer I (e.g. standard greetings, farewells, and introductions) or IV (phrasing answers given the knowledge in the brain and the goals defined in layer II). The robot's responses are produced using a set of rules to transform an RDF triple into English language. For English, we  created a grammar using the concepts \textit{person} (first, second, or third person pronouns or names), \textit{object}, \textit{location} and lists of \textit{properties}. With this basic grammar, the robot can already understand and generate a large portion of common language. In the future, we will use WordNet to produce more varied responses and extend the grammar to capture more varied input and roles.

\item \textbf{Knowledge Representation}
% a)the areas we cover currently, b) linking to other ontologies and query of dbpedia, and c) the power on inferencing.
The robot's brain must store and represent knowledge about the world, and the perspectives about it. For the latter we use the GRaSP ontology, as mentioned in Section \ref{sec:grasp}. For the former, we created our own ontology ``Nice to meet You", which covers the basic concepts and relations for human-robot social interaction (e.g. a person's name, place of origin, occupation, interests). Our ontology complies with 5 Star Linked Open Data, and is linked to vocabularies like FOAF and schema.org. Furthermore, the brain is able to query factual services like Wolfram Alpha, and LOD resources like DBpedia and GeoNames. The robot's brain is hosted in a GraphDB triple store. Given the above, this layer allows for two main interactions with the brain. The first is to process a statement, which implies generating and uploading the  corresponding triples to the brain with source and perspective values. The second is to process a question, where a SPARQL query is fired against the brain. The result, being an empty list or a list with one or more results, is passed to layer III to generate a response. A list of values may represent conflicting information (disjunctive values) or multiple values (orthogonal values), each triggering different responses. In the future, we will extend the capabilities by enabling the robot to reason over its knowledge and generate new knowledge through inferencing. 
\end{enumerate}

\vspace{-15pt}
\section{Conclusions and Future Work}
\label{sec:future_work}
\vspace{-10pt}

We described a model for a robot's world perception and communication, using an RDF repository modeling a theory of mind. It allows us to deal with uncertainties, conflicts, gaps, and different perspectives on this world. The state of information and knowledge the robot obtains can drive  communication, e.g. the desire to resolve conflicts or uncertainties. Currently, the scripts and knowledge are limited. For the future, we plan to include different conversational models, NLP modules and types of knowledge. Furthermore, we extend the platform to empirically evaluate performance. Finally, we make the system functional through task-based dialogues and the processing knowledge and situation-aware. Our implementation is available as open source: \url{https://github.com/cltl/pepper}.

\vspace{-10pt}
\section*{Acknowledgement}
\vspace{-10pt}
This research was funded by the VU University Amsterdam and the Netherlands Organization for Scientific Research via the Spinoza grant awarded to Piek Vossen. We also thank Bob van der Graft for his support.

%\newpage
\medskip
\bibliographystyle{splncs04}
\bibliography{leolani}

\twocolumn

\section*{Appendix: Dialogues}
{\scriptsize In the dialogues, \textit{L} preceding an utterance stands for Leolani, other letters preceding utterances stand for various people. Perceptions of the robot of people and objects are marked using square brackets, e.g. [Sees a new face].}

%\vspace{-15pt}
%\begin{table*}
%\begin{tabular}{L{14cm}}
%In the dialogues, \textit{L} preceding an utterance stands for Leolani, other letters preceding utterances stand for various people. Perceptions of the robot of people and objects are marked using square brackets, e.g. [Sees a new face].
%\end{tabular}
%\end{table*}
%\vspace{-15pt}

\vspace{-15pt}
\begin{table}[!ht]
{\scriptsize
\begin{tabular}{L{0.5cm}L{5cm}} 
\multicolumn{2}{l} {Meeting a new person}\\
\hline
L: & [Sees the face of an unknown woman]\\
L: &Hi there, I would like to know you.\\
L: &My name is Leolani, what is your name?\\
H: & My name is Selene.\\
L: &[Stores the triples to represent Selene as a new friend. If the confidence for the name is too low, asks for confirmation]\\
L: &I hope I am correct and your name is: Selene. \\
H:& Yes that is my name.\\
L: &Nice to meet you Selene. Now I have a new friend.\\
L: &[Queries the brain for information on Selene to fill social gaps]\\
L: &Where are you from?\\
H: &I am from Mexico.\\
L: &[Stores the triples and queries the brain for friends from Mexico. The query just returns Selene.]\\
L: &Now I know 1 person from Mexico.\\
\hline
\end{tabular}
\label{exa:get-to-know}
%\caption{Leolani (L)}
}
\end{table}
\vspace{-15pt}

\vspace{-15pt}
\begin{table}[!ht]
{\scriptsize
\begin{tabular}{L{0.5cm}L{5cm}}  
\multicolumn{2}{l} {Handling conflicting information}\\
\hline
L: & [Sees a human face and recognizes Lenka]\\
L: & Hi Lenka, nice to see you.\\
L: & [Queries the brain and discovers she does not know  where Lenka is from]\\
L: &Lenka, where are you from?\\
H: &I am from Serbia.\\
L: &[Stores the triples and queries the brain for friends from Serbia and finds only Lenka.]\\
L: &Nice, I did not know anybody from Serbia yet.\\
H: &Where is Bram from?\\
L: &[Queries the brain for origin of Bram. Finds Bram as a friend and information where he is from.]\\
L: &Bram is from the Netherlands.\\
H: &Bram likes romantic movies.\\
L: &[Stores the triples with Lenka as source.]\\
L: &You told me that Bram likes romantic movies.\\
L: & [Sees Lenka leave and sees Bram arrive.]\\
L: &Hi Bram.\\
B: &I like science fiction movies.\\
L: &[Stores the triples and queries what Bram likes. Gets more than one value while expecting one.]\\
L: &I am surprised.\\
L: &Bram likes romantic movies, says Lenka.\\
L: &Bram likes science fiction movies, says Bram.\\
%L: &humans are confusing.\\
\end{tabular}
\label{exa:conflict2}
%\caption{}
}
\end{table}
\vspace{-15pt}

\vspace{-15pt}
\begin{table}[!ht]
{\scriptsize
\begin{tabular}{L{0.5cm}L{5cm}} 
\multicolumn{2}{l} {Checking information and trust}\\ 
\hline
L: & [Sees a human face and recognizes Bram]\\
L: &Greetings Bram. Nice to see you again.\\
B: &Do you know where I am from?\\
L: &You are from the Netherlands, you said.\\
B: &Do you also know Lenka?\\
L: &Yes I know her, she is a very good friend of mine.\\
B: & Where is she from?\\
L: & Lenka is from Serbia, Lenka said\\
B: & Do you believe Lenka?\\
L:& [Queries the brain for perspective information on Lenka making this statement. Finds no uncertainty or denial.]\\
L: &I believe her.\\
\end{tabular}
\label{exa:conflict3}
%\caption{}
}
\end{table}
\vspace{-15pt}

\vspace{-30pt}
\begin{table}[!ht]
{\scriptsize
\begin{tabular}{L{0.5cm}L{5cm}} 
\multicolumn{2}{l} {Observing the environment}\\ 
\hline
L: & [Sees an object and detects it as a cat with certain confidence]\\
L: & [Shortly after that she sees a human face and recognizes Bram]\\
L: &Greetings Bram. Nice to see you again.\\
L: & Guess what? I just saw a cat!\\
B: & That is not a cat but a rabbit.\\
L: & [Labels the image previously recognized as a cat instead as a rabbit and stores it in her brain]\\
B: & I like this rabbit.\\
L: & [Sees Bram leave and sees Selene arrive]\\
L: & Hi Selene. Greetings.\\
L: & [The presence of the rabbit is prominent, if the object is seen again recognition of a rabbit overrides recognition of a cat despite a higher score of the latter]\\
L: & Guess what, I just met a rabbit.\\
S: & A rabbit bites.\\
S: & I like a cat more.\\
S: & Have you ever seen a cat?\\
L: & [Queries the brain for objects labeled as cat and perceived by the sensor]\\
L: & No I have never seen a cat.\\
S: & What animals did you see?\\
L: & [Queries the brain for objects labeled as subtype of animal and perceived by the sensor]\\
L: & I saw a rabbit and a panda.\\
S: & What does rabbit do?\\
L: & [Queries the brain for properties assigned to the rabbit through conversations]\\
L: & Rabbits bite, Selene said.\\
L: & Rabbits cuddle, Bram said.\\
S: & Who likes rabbits?\\
L: & [Queries the brain for like perspective on the rabbit through conversations]\\
L: & Bram likes rabbits, Bram said.\\
\end{tabular}
\label{exa:conflict-vision}
%\caption{}
}
\end{table}
\vspace{-15pt}

% Not very interesting
%\vspace{-15pt}
%\begin{table}[!ht]
%{\scriptsize
%\begin{tabular}{L{0.5cm}L{5cm}} 
%\multicolumn{2}{l} {Asking for her friends}\\
%\hline
%L: &Greetings Selene. Nice to see you again?\\
%S: &I was wondering?\\
%S: &Who are your friends?\\
%L: &Gives the list\\
%S: &Goodbye\\
%L: &See you soon\\
%\end{tabular}
%\label{exa:list-of-all-friends}
%%\caption{}
%}
%\end{table}
%\vspace{-15pt}

% Not very interesting
%\vspace{-15pt}
%\begin{table}[!ht]
%{\scriptsize
%\begin{tabular}{L{0.5cm}L{5cm}} 
%\multicolumn{2}{l} {Asking for what people like}\\ 
%\hline
%S: & Hi Leolani\\
%L: &Hi Selene\\
%S: &I like ?deep learning?\\
%S: &I am from ?Ecuador?\\
%S: &Piek likes ?romantic movies?\\
%S: &You like electricity\\
%S: &What do you like?\\
%L: &list of things Leolani likes?.\\
%\end{tabular}
%\label{exa:likes}
%%\caption{}
%}
%\end{table}
%\vspace{-15pt}

% Not very interesting
%\vspace{-15pt}
%\begin{table}[!ht]
%{\scriptsize
%\begin{tabular}{L{0.5cm}L{5cm}}  
%\multicolumn{2}{l} {Asking for what people like}\\ 
%\hline
%L: &Hi Piek\\
%P: &I like ?horror movies?\\
%P: &What do I like?\\
%L: &list of things with the names of the sources what Piek likes\\
%\end{tabular}
%\label{exa:horror-and-romantic-movies}
%%\caption{}
%}
%\end{table}
%\vspace{-15pt}

%\vspace{-15pt}
%\begin{table}[!ht]
%{\scriptsize
%\begin{tabular}{L{0.5cm}L{5cm}} 
%\multicolumn{2}{l} {Asking for a dump of her brain}\\  
%\hline
%S: &How many things do you know?\\
%S: &Who have you talked to?\\
%S: &How many friends are from Amsterdam?\\
%S: &Tell me everything I told you.\\
%\end{tabular}
%\label{exa:brain-dump}
%\caption{}
%}
%\end{table}
%\vspace{-15pt}

\end{document}